\pdfoutput=1

\documentclass[11pt]{article}

\usepackage[preprint]{acl}

\usepackage{times}
\usepackage{latexsym}

\usepackage{geometry}
\usepackage{multirow}
\usepackage{booktabs}
\usepackage{array}
\usepackage{caption}
\usepackage{tabularx}
\usepackage{arydshln}

\usepackage[T1]{fontenc}

\usepackage[utf8]{inputenc}

\usepackage{microtype}

\usepackage{inconsolata}

\usepackage{graphicx}

\title{Explore the Reasoning Capability of LLMs in the Chess Testbed}

\author{
\quad
 \textbf{Shu Wang\textsuperscript{1}},
 \textbf{Lei Ji\textsuperscript{2}},
 \textbf{Renxi Wang\textsuperscript{3}},
 \textbf{Wenxiao Zhao\textsuperscript{1}},
 \textbf{Haokun Liu\textsuperscript{4}},
 \textbf{Yifan Hou\textsuperscript{5}},
 \textbf{Ying Nian Wu\textsuperscript{1}}\\
\\
 \textsuperscript{1}UCLA,
 \textsuperscript{2}Microsoft Research,
 \textsuperscript{3}MBZUAI,
 \textsuperscript{4}University of Toronto,
 \textsuperscript{5}Peking University
}
\begin{document}
\maketitle
\begin{abstract}

Reasoning is a central capability of human intelligence. In recent years, with the advent of large-scale datasets, pretrained large language models have emerged with new capabilities, including reasoning. However, these models still struggle with long-term, complex reasoning tasks, such as playing chess. Based on the observation that expert chess players employ a dual approach combining long-term strategic play with short-term tactical play along with language explanation, we propose improving the reasoning capability of large language models in chess by integrating annotated strategy and tactic. Specifically, we collect a dataset named MATE\footnote{ \url{https://mate-chess.github.io/}\quad\quad\quad\quad\quad\quad\quad
Correspondence to: Shu Wang<shuwang0712@ucla.edu>.
Yifan Hou is a four-time chess world champion. 
}, which consists of 1 million chess positions with candidate moves annotated by chess experts for strategy and tactics. We finetune the LLaMA-3-8B model and compare it against state-of-the-art commercial language models in the task of selecting better chess moves. Our experiments show that our models perform better than GPT, Claude, and Gemini models. We find that language explanations can enhance the reasoning capability of large language models.
\end{abstract}

\section{Introduction}
\begin{quote}
``Strategy without tactics is the slowest route to victory. Tactics without strategy is the noise before defeat.''  ----Sun Tzu
\end{quote}

Rational thought and deliberate cognition rely heavily on reasoning, a core component of human intelligence\cite{garnham1994thinking}. Given sufficient information, people can logically progress through a sequence of steps. In the field of artificial intelligence\cite{russell2016artificial}, it has been a persistent objective to study the reasoning capability, as it is essential for both problem-solving and decision-making processes.

The past few years have seen large language models exhibit extraordinary aptitude in the tasks that require reasoning capability\cite{brown2020language,wei2022emergent, kojima2022large,bubeck2023sparks}. However, language models show significant limitations in planning and reasoning for complicated tasks\cite{xu2023large, dziri2024faith, srivastava2022beyond, wang2024llm, mirzadeh2024gsm}. In this paper, we use chess as a testbed to study how we can improve the reasoning capability of large language models for complex tasks.

\begin{figure}[t!]
    \centering
    \includegraphics[width=\linewidth]{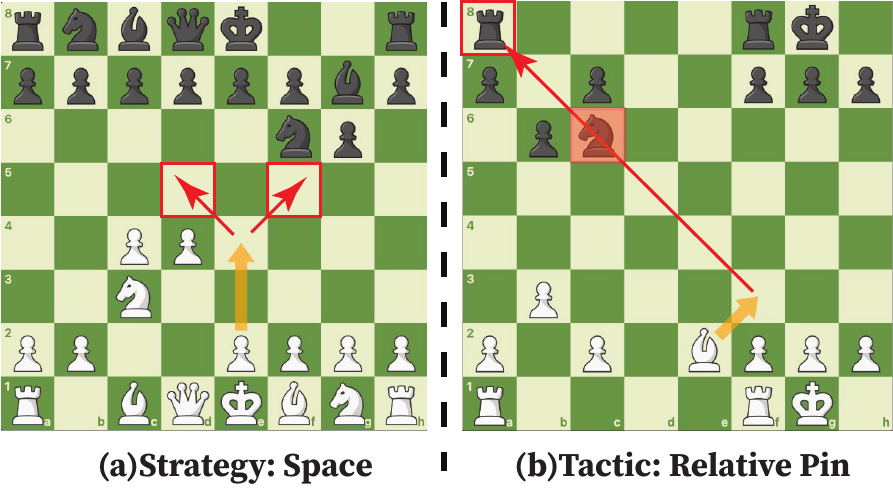}
    \caption{\textbf{Strategy and Tactic} (a)White E2 pawn moves to E4, takes more space in the center, and exerts pressure on black. Black will have a hard time struggling to develop its pieces. (b)White E2 bishop moves to F3 and pins the knight on C6. The black knight cannot move, or the A8 rook behind the knight will be taken. White will take black knight for free in the next move.}
    \label{fig:teaser}
\end{figure}

 Chess reasoning is challenging, requiring analytical calculation and intuitive insights. Good chess players employ a dual approach, which includes (i) Long-term Strategy: It relies on rapid, intuitive thinking based on the pattern recognition of the chess board. (ii) Short-term Tactic: It involves slow, analytic calculations that typically consider 1-6 moves ahead, depending on the player's skill level. Figure \ref{fig:teaser} shows an example of strategy and tactic. Notably, experienced players think out loud: they develop strategic plans in clear language, and they evaluate the afterward position in lucid words after calculating the precise moves of a tactic.

Drawing inspiration from the thinking approach used by chess experts, we propose a method to enhance large language models' chess-playing capabilities by incorporating both strategy and tactic in language annotation. We collect the MATE(Move on strAtegy and Tactics datasEt), a dataset of around 1 million chess positions, and annotate the candidate moves for each position with long-term strategy and short-term tactic. Then, we utilize the MATE to finetune open source large language models. Finally, we evaluate the performance of our models and compare them against state-of-the-art large language models. Our models outperform the best commercial language model by 24.2\% when both strategy and tactic are provided. 

In summary, this work's contributions are three-fold:
\begin{itemize}
    \item We collect a high-quality chess dataset. For each position, the candidate moves are provided with a description of the strategy and tactic information annotated by experienced chess players, including world champion-level experts.
    \item We find that language explanations can enhance the reasoning capability of large language models.
    \item We discover that integrating the dual-mode of strategy and tactic can improve the chess-playing capability of language models.
\end{itemize}

\section{Related Work}

Chess has historically been esteemed as a challenging intellectual pursuit\cite{thrun1994learning}. With all the rules and the chess board provided, it is a pure reasoning task without any uncertainty or randomness. In 1997, Deep Blue, created by IBM, defeated the chess world champion—Russian player Garry Kasparov—in a match that astonished the world. Modern chess engines such as Stockfish, AlphaZero\cite{silver2017mastering}, Leela Chess Zero, which integrate search algorithms, deep neural networks, and reinforcement learning, play significantly better than the strongest human players. Recent work\cite{ruoss2024grandmaster} trains a transformer model on millions of annotated chess games, enabling it to play precise and beautiful chess.

Though chess is a “solved problem” in the field of artificial intelligence, many researchers used it as a testbed to study the capabilities of language models\cite{kamlish2019sentimate, noever2020chess, toshniwal2022chess, deleo2022learning, alrdahi2023learning}. Large language models have demonstrated remarkable capabilities across a diverse range of tasks\cite{li2024graph, wang2024toolgen, jiang2024raising}, and \cite {fauber2024learning} shows by instruction fine-tuning, language models can learn how to move a pawn or a piece legally. \citet{feng2024chessgpt} collects a dataset of chess games and chess-related corpus, then trains language models capable of effectively tracking chess board states. \citet{guo2024can} consider large language models as the action space pruner and the value function approximator, boosting the Monte-Carlo Tree Search algorithm for playing chess. Unlike other works, our research focuses on whether strategic and tactical explanations can guide language models to find better moves.

\section{MATE}
\begin{figure}[t!]
    \centering
    \includegraphics[width=\linewidth]{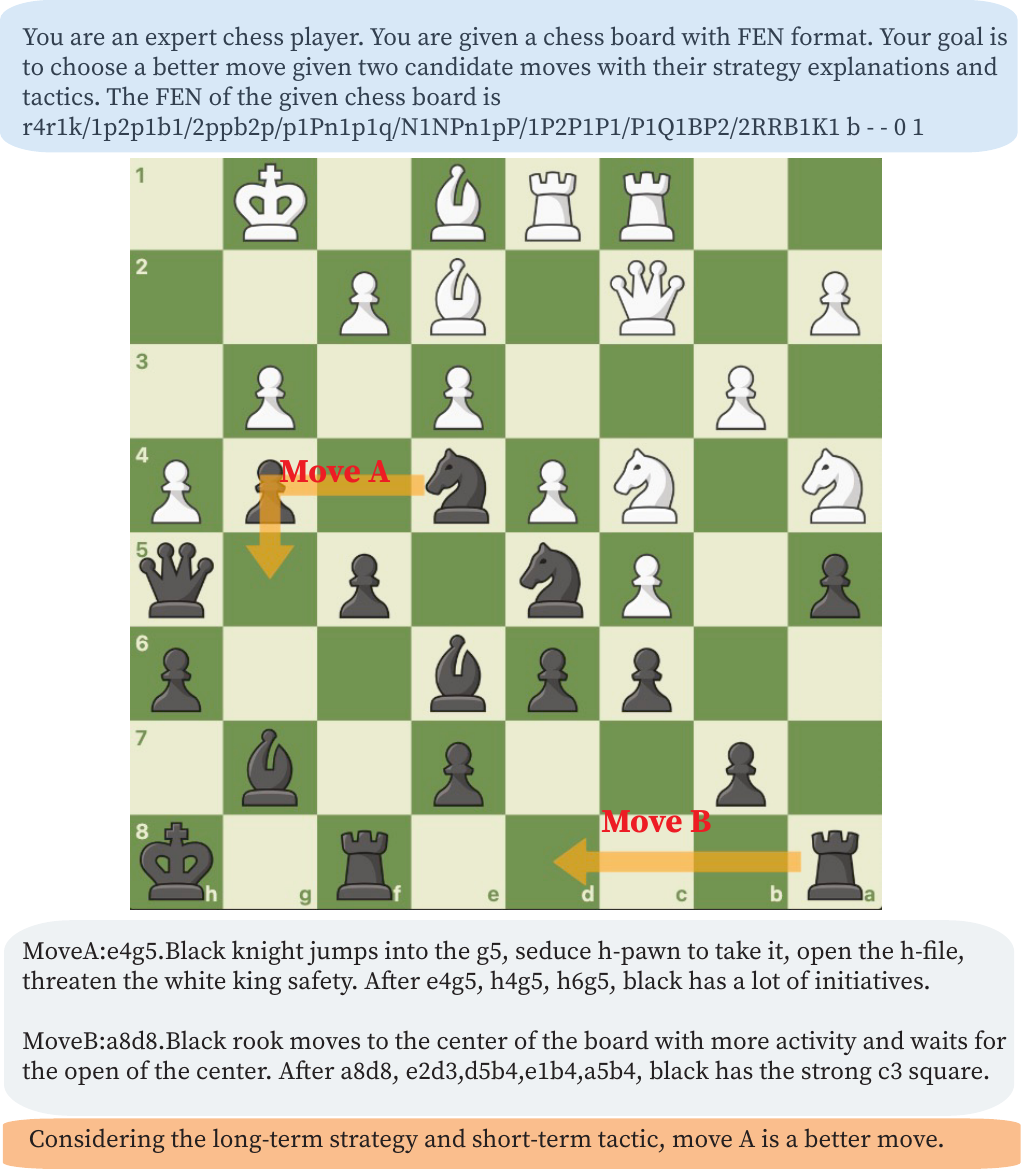}
    \caption{A data example in MATE-Strategy\&Tactic.}
    \label{fig:example}
\end{figure}

We propose the MATE(Move on strAtegy and Tactic datasEt) for exploring the reasoning capability of large language models in chess. In chess, mate is known as checkmate, which occurs when a king is placed in check and has no legal moves to escape. Checkmating the opponent wins the game.

We collect around 1 million chess positions from the open source chess server -- Lichess. The data collection guidelines can be found in Appendix \ref{appendix:data}. The positions are either selected from chess games or chess puzzles. These specific board positions ask players to play moves to achieve a particular goal, such as checkmating or gaining a material advantage. Analyzing these positions can be an efficient method to enhance chess skills without committing to full games. We use the Forsyth-Edwards Notation(FEN) format to describe the board position. FEN is a notation in one line of text with only ASCII characters(Appendix \ref{appendix:notation}).

For each position, we select multiple reasonable moves and then annotate each move with language explanations of long-term strategy and short-term tactic by expert chess players. We use the Universal Chess Interface(UCI) format to denote the move. For a specific move, UCI encodes the start and end squares of that pawn or piece.

For chess strategy annotation, we categorize the future strategical plan into five kinds: (i) material count, (ii) piece activity, (iii) pawn structure, (iv) space, and (v) king safety. We ask chess experts, including world champion-level players, to
formulate the rules to determine the optimal strategy for any position(Appendix \ref{appendix:strategy}). For each strategic category, there are approximately 20 distinct linguistic expressions to describe the corresponding plan. 

For chess tactic annotation, the multitude of categories is overwhelming(Appendix \ref{appendix:tactic}): skewer, pin, fork, x-ray, remove the defender, overload, Greek gift, windmill, discovered attack, inflection, etc. For simplicity,  we list the sequence of moves and provide a factual description of the resulting position. Unlike search algorithms that explore long tactical reasoning chains, our approach focuses on short-term calculations, limiting the move sequence length. The move sequences are generated using the open source chess engine Stockfish.  

We evaluate move quality using Stockfish, assigning a hidden score to each move. In our dataset, we select two moves for each position whose differences in scores exceed a specified threshold. This significant score gap clearly indicates one move is superior to the other.

We create four sub-dataset based on the MATE: (i) MATE-No-Explanation: given chess positions, the candidate moves are provided without strategical nor tactical explanation; (ii) MATE-Strategy: given chess positions, the candidate moves are provided with strategical elaboration; (iii) MATE-Tactic: given chess positions, candidate moves are provided with tactical description; (iv) MATE-Strategy\&Tactic: given chess positions, candidate moves are provided with both strategy and tactic, a sample is shown in Figure \ref{fig:example}. We investigate the difficulty levels of positions for each sub-dataset and find they are at similar levels.

\begin{figure}[t!]
    \centering
    \includegraphics[width=\linewidth]{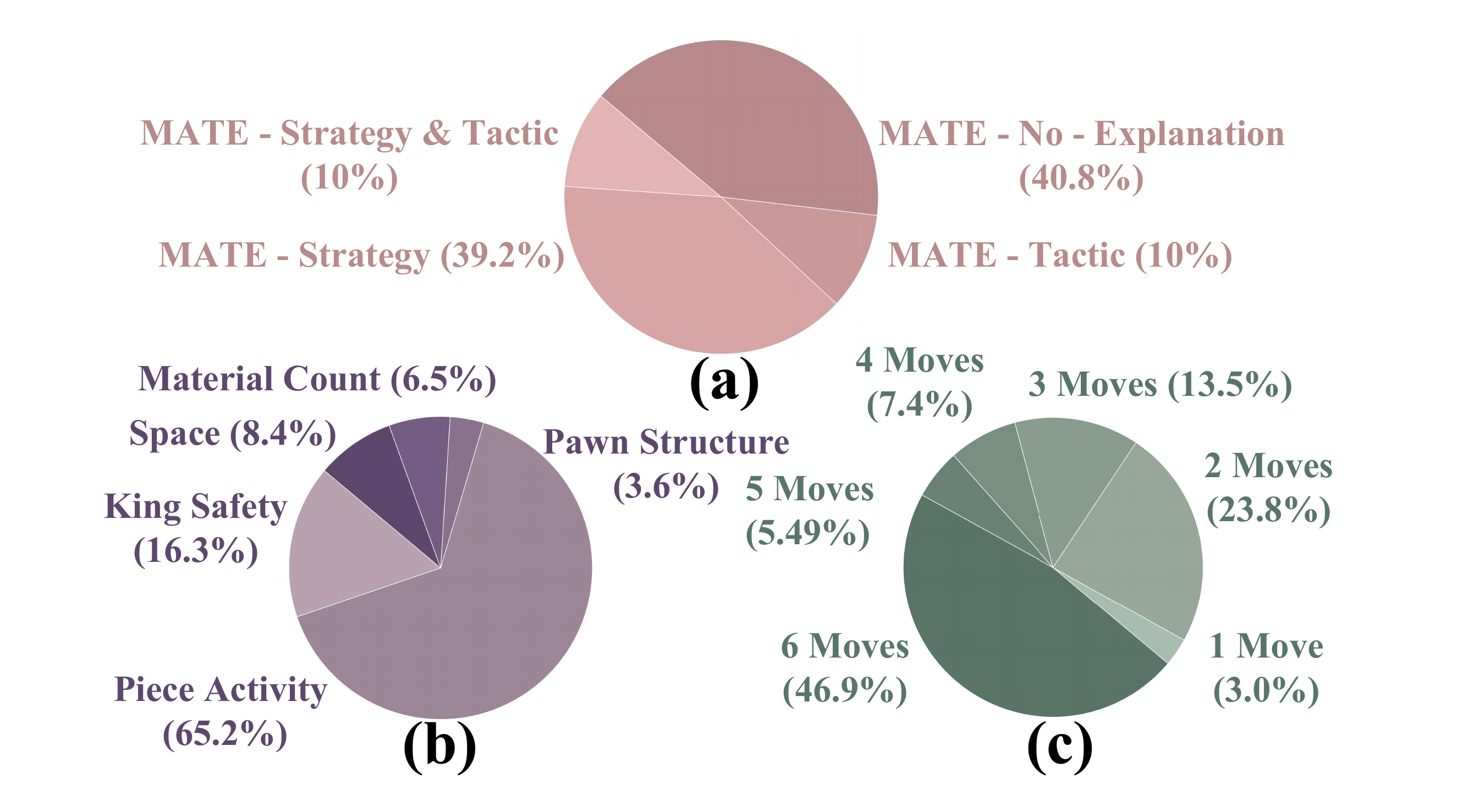}
    \caption{\textbf{Dataset Summary} (a)Distribution of samples across the MATE subsets. (b)Distribution of strategy in the MATE. (c)Distribution of tactic in the MATE.}
    \label{fig:summary}
\end{figure}

Most positions in the MATE lend themselves to long-term strategic planning. While many positions are generally not very sharp, meaning there are no immediate opportunities to gain an advantage through tactical play, we can still formulate strategic plans for them. Consequently, we are unable to identify short-term tactics for these positions. As a result, the MATE-Strategy subset is significantly larger than both the MATE-tactic and MATE-Strategy\&Tactic subsets.
We show the summary of the MATE in Figure \ref{fig:summary}.

\section{Experiments}
\begin{table*}[h]
\centering
\begin{tabularx}{\textwidth}{>{\centering\arraybackslash}p{3cm} *{4}{>{\centering\arraybackslash}X} *{4}{>{\centering\arraybackslash}X}}
\toprule
\multirow{2}{*}{\textbf{Model}} & \multicolumn{4}{c}{\textbf{Zero-Shot Setting}} & \multicolumn{4}{c}{\textbf{Few-Shot Setting}} \\
\cmidrule(lr){2-5} \cmidrule(lr){6-9}
& {\small\textbf{N}} & {\small\textbf{S}} & {\small\textbf{T}} & {\small\textbf{ST}} & {\small\textbf{N}} & {\small\textbf{S}} & {\small\textbf{T}} & {\small\textbf{ST}} \\
\midrule
\small{gpt-4} & 53.1 & 54.6 & 60.0 & 60.0 & 54.7 & 58.9 & 57.7 & 68.1 \\
\small{gpt-4o} & 46.4 & 52.8 & 54.8 & 60.1 & 48.5 & 54.3 & 52.7 & 63.1 \\
\small{o1-mini} & 51.5 & 58.8 & 64.1 & 69.2 & 50.4 & 58.3 & 62.0 & 65.9 \\
\small{o1-preview*} & \underline{56.4} & \underline{65.4} & \underline{77.2} & \underline{76.6} & \underline{59.0} & \underline{65.4} & \underline{76.2} & \underline{78.6} \\
\small{claude-3.5-sonnet} & 49.6 & 54.9 & 56.9 & 54.9 & 51.9 & 63.7 & 59.9 & 66.1 \\
\small{claude-3-opus} & 48.3 & 54.5 & 53.7 & 57.3 & 51.0 & 55.8 & 53.2 & 60.2 \\
\small{gemini-1.5-pro} & 50.6 & 48.8 & 54.2 & 52.6 & 50.5 & 50.1 & 52.7 & 50.4 \\
\small{gemini-1.5-flash} & 46.1 & 50.8 & 54.2 & 52.9 & 49.7 & 48.2 & 53.8 & 55.6 \\
\small{Ours-no-explanation} & \textbf{63.5} & -- & -- & -- & \textbf{64.7} & -- & -- & -- \\
\small{Ours-strategy} & -- & \textbf{89.7} & -- & -- & -- & \textbf{89.8} & -- & -- \\
\small{Ours-tactic} & -- & -- & \textbf{94.6} & -- & -- & -- & \textbf{94.5} & -- \\
\small{Ours-strategy\&tactic} & -- & -- & -- & \textbf{95.2} & -- & -- & -- & \textbf{95.3} \\

\bottomrule
\end{tabularx}

\caption{Experimental results in terms of accuracy(\%) on MATE. The best-performing score is highlighted in \textbf{bold}, and the second-best is \underline{underlined}. In the table, N stands for MATE-N, S stands for MATE-S, T stands for MATE-T, and ST stands for MATE-ST.} 
\label{table1}
\end{table*}

\subsection{Experiment Setup}

We train our models using the pretrained Llama-3-8B model\cite{dubey2024llama} as the foundation. The models are finetuned with llamafactory\cite{zheng2024llamafactory}, employing a cosine learning rate scheduler with 3\% warm-up steps. We set the maximum learning rate to $5\times 10^{-6}$. We use DeepSpeed ZeRO Stage 3 \cite{rajbhandari2020zero} across 4$\times$H100 GPUs. We train the models for 5 epochs.

We incorporate specific tokens in FEN format to enhance the foundation model's understanding of chessboard positions. We add the <line> token to separate each row of the board and the <color> token to indicate which side is to move next. Our experiments show no significant difference in performance with or without these special tokens.

We train four models with MATE-No-Explanation(MATE-N), MATE-Strategy(MATE-S), MATE-Tactic(MATE-T), and MATE-Strategy\&Tactic(MATE-ST), respectively. 

We compare our models with the following baselines:
\begin{itemize}
    \item  GPT: gpt-4-0613, gpt-4o-2024-08-06, o1-preview-2024-09-12, o1-mini-2024-09-12;
    \item  Claude: claude-3.5-sonnet, claude-3-opus;
    \item  Gemini: gemini-1.5-pro, gemini-1.5-flash.
\end{itemize} 

In our experiment, we have the zero-shot setting and the few-shot setting. In the zero-shot  setting, models are evaluated on their inherent reasoning capabilities without any prior examples. In the few-shot setting, a few examples are given
to the models before the test example. We evaluate models on 1000 samples in the individual test sets for each setting. In each test sample, models score when they output the optimal move from candidate moves.

\subsection{Results}

Our experimental results in Table \ref{table1} shows:
(i) MATE proves sufficiently complex to differentiate among commercial LLMs. Our results demonstrate that the o1-preview model leads in performance by a substantial margin. 
(ii)Interestingly, prompting strategies do not significantly impact performance in our task. We observe no substantial improvement in performance when adopting a few-shot setting compared to a zero-shot setting. (iii)Our models exhibit superior reasoning capabilities compared to commercial models, as demonstrated by their performance across various test sets.


\textbf{Language enhances chess-reasoning in language models.} While some researchers argue that language is not used for reasoning\cite{fedorenko2024language}, our findings lead us to a contradictory conclusion in chess. Our evaluations demonstrate that performance improves for most LLMs we test when provided with linguistic explanations. Using o1-mini in the zero-shot setting as an example, its performance improved by 14\% on the MATE-S, 24\% on the MATE-T, and 34\% on the MATE-ST, all compared to its baseline performance on the MATE-N.

\textbf{Integrating long-term strategy and short-term tactics enhances language models' chess-playing ability.} Most models demonstrate superior performance in the MATE-ST subset compared to other subsets. For instance, gpt-4o demonstrates the following improvements in the MATE-ST zero-shot setting: a 10\% increase compared to MATE-T, a 14\% increase compared to MATE-S, and a 30\% improvement relative to MATE-N.

We conduct additional experiments to evaluate: (1) model performance with multiple candidate moves, (2) the quality of strategy explanations generated by both our trained models and commercial models, and (3) the difficulty levels of chess positions across sub-datasets, assessed through both human evaluation and language models' evaluation. The details of additional experiments can be found in Appendix \ref{appendix:multimove}, \ref{appendix:generate}, and \ref{appendix:difficulty}.

In future, the combination of long-term strategic planning and short-term tactical decision-making can be applied to strengthen language models' reasoning capabilities across various tasks.

\section{Conclusion}
We propose a method to enhance LLMs' chess-reasoning capabilities by incorporating strategy and tactic annotations. We craft the MATE, train our models and compare them against state-of-the-art commercial language models. Our models outperform others in the chess-reasoning task. We find language helps language models' reasoning. We demonstrate combining long-term intuition with short-term analysis can be a promising direction for exploration.

\section*{Acknowledgment}
We thank Dr.Pan Lu, Dr.Wenhu Chen and Han Jiang for fruitful discussions. Y. W. was partially supported by NSF DMS-2015577, NSF DMS-2415226, and a gift fund from Amazon.

\section*{Limitation}
Although the idea of combining strategy and tactics is prevalent in all games, we only study chess. A comprehensive study of multiple game types should demonstrate this approach's effect better. 

We use chess puzzles to test the models' ability, asking the model to choose between two plausible moves. This is a common way for professional players to exercise. However, the ideal scenario would require running a complete game on the chess engine to test a model's full strength and ability to carry out strategy and tactics. 

Our dataset is annotated by chess experts. However, we acknowledge that potential biases may exist in determining appropriate strategies for various positions and in evaluating post-tactical situations. Furthermore, the limited number of chess experts may only capture the thought processes of a subset of all players.

Our experiment only uses LLaMA-3-8B for fine-tuning, so we don't understand how the improvement changes to model sizes and base model quality.

\bibliography{custom}

\appendix

\section{Appendix}
\subsection{Data Collection Guidelines}\label{appendix:data}
In order to represent the full characteristics of chess games, our dataset adheres to the following collection guidelines:

(1)it covers all phases of a chess game, including openning, middlegame, endgame;

(2)it involves different strategies and tactics;

(3)it origins from different levels of chess players' games and different difficulty level of puzzles.

\subsection{Chess Notation}\label{appendix:notation}
\paragraph{FEN}
Forsyth-Edwards Notation, abbreviated as FEN, is the standard method for describing chess positions. This system was developed by Steven J. Edwards, a computer programmer, who adapted an earlier notation created by journalist David Forsyth. Edwards' modifications made the notation compatible with chess software, enhancing its utility in the digital age.

FEN encodes chess positions using the following elements:(1)
Piece positions: Capital letters for white pieces, lowercase for black. Numbers indicate empty squares.
(2) Active color: w for white's turn, b for black's.
(3) Castling rights: K means white kingside, Q means white queenside, k means black kingside, q means black queenside.
(4) En passant target square: If a pawn has just moved two squares, this is the square behind it.
(5) Halfmove clock: Moves since the last pawn advance or capture.
(6) Fullmove number: The number of completed turns in the game.

Board rows are separated by forward slashes /. This compact notation allows for precise representation of any chess position, facilitating analysis and game reconstruction.

\paragraph{UCI}The Universal Chess Interface is an open communication protocol that facilitates interaction between chess engines and user interfaces. UCI encodes chess moves using a four-character system that represents the starting and ending coordinates of a piece's movement. Each move is denoted by a combination of two letters and two digits, such as "e2e4", which indicates moving a piece or a pawn from square e2 to e4.

\subsection{Chess Strategy}\label{appendix:strategy}

We elaborate on the details of each strategy, including the criteria we use to identify them.

\paragraph{Material Count}It is a fundamental strategy, particularly for beginners. While the game ultimately aims for checkmate, having a material advantage often influences the result more frequently. Each piece is assigned a specific value, and understanding these values helps players assess their position. When other elements are relatively equal, prioritizing material acquisition can lead to a decisive advantage in the game. This strategy is most relevant when there is an imbalance in material comparison and both kings are safe. It generally applies to most types of positions, though king safety may occasionally take precedence.

\paragraph{Piece Activity}It is an advanced strategy, focuses on the placement and effectiveness of pieces rather than just their assigned value. In some situations, players may have an equal material count, but the effectiveness of their pieces can vary significantly. Pieces positioned centrally are typically more powerful, allowing for greater control and flexibility. This strategy is especially relevant in dynamic positions where the mobility of pieces can lead to tactical opportunities. Focus on piece activity when there is a marked difference in piece positioning, such as when some pieces occupy central squares while others remain in the corners. This is especially crucial in dynamic positions, particularly when one side is attacking.

\paragraph{Space}Gaining a spatial advantage is closely related to piece activity and can greatly impact a player's effectiveness. When one side controls more space on the board, their pieces can move more freely and exert influence over critical areas. This advantage can limit the opponent's options and create opportunities for attack. Space is a vital evaluation factor, particularly in positional play, where controlling key squares can lead to long-term advantages. Space advantage typically arises in the opening and middlegame, especially when more pawns are on the board, as this can enhance spatial control.

\paragraph{Pawn Structure}The configuration of pawns is a unique and complex aspect of chess strategy. With eight pawns per side, the formation can vary widely, influencing both positional and dynamic play. Strong pawn structures can create weaknesses for the opponent, while poorly positioned pawns can become liabilities. Understanding pawn dynamics is essential for developing long-term strategies and can dictate the overall flow of the game. Consider pawn structure when faced with clear issues such as doubled or isolated pawns. Typical positions arising from certain openings, like the Sicilian or Ruy Lopez, should also prompt a focus on pawn structure.

\paragraph{King Safety}Ensuring king safety is a critical strategy throughout the game. A secure king allows other strategies to be executed more effectively, while a vulnerable king can lead to immediate threats and checkmate. Prioritizing king safety not only protects against attacks but also enables players to focus on their offensive strategies with confidence. This strategy should always be considered alongside the others to maintain a balanced approach to the game. Assess king safety when the king is exposed, particularly without pawns in front of it, and when the opponent's pieces are coordinated to attack, possibly leveraging tactical combinations along open files.

\subsection{Chess Tactic}\label{appendix:tactic}
Here we list several common tactics in chess:
\paragraph{Pin}Pin tactics occur when an attacked piece cannot move without exposing an even more valuable piece (or target) behind it.
\paragraph{Fork}A fork is a type of double attack whereby a single piece makes multiple threats.
\paragraph{Battery}In chess, a battery refers to lining up two or more pieces on the same diagonal, rank or file. Only queens, rooks and bishops can form a battery. The rooks can form a battery on a rank or file whilst the bishops can be part of a battery on a diagonal. The queen, of course, can be part of a battery on a rank, file or diagonal.
\paragraph{X-Ray}X-Ray refers to the ability of long-range pieces to see “through” an enemy piece. This tactical idea is sometimes referred to as an x-ray attack, but it can also be used as a defensive tactic.
\paragraph{Discovered Attack}A discovered attack occurs when moving a piece reveals a strong threat from a piece hiding behind it. The power of a discovered attack often lies in the fact that you can use it to set up a double attack.
\paragraph{Windmill}A windmill tactic can also be described as a series of forced discovered attacks. This tactic is also known as a see-saw, based on how the front piece keeps returning to its previous position. 
\paragraph{Greek Gift}The Greek Gift Sacrifice (also known as the classical bishop sacrifice) is a specific case of demolition of the pawn structure in front of the enemy king. A key feature of the Greek Gift Sacrifice is the placement of the white bishop on d3, the white knight on f3 and the white queen on d1, all ready to join in the attack against black’s king
\paragraph{Double Attack}A double attack is a situation where one or more of your pieces make multiple threats. A double attack performed by a single piece is known as a fork.

\subsection{Experiments on Multiple Candidate Moves}\label{appendix:multimove}
\begin{table}[htbp]
\centering
\begin{tabularx}{\columnwidth}{>{\centering\arraybackslash}p{3cm} *{4}{>{\centering\arraybackslash}X}}
\toprule
\multirow{2}{*}{\textbf{Model}} & \multicolumn{4}{c}{\textbf{Zero-Shot Setting}} \\
\cmidrule(lr){2-5}
& {\small\textbf{N}} & {\small\textbf{S}} & {\small\textbf{T}} & {\small\textbf{ST}} \\
\midrule
\small{gpt-4} & 37.4 & 40.1 & 61.7 & 56.3 \\
\small{gpt-4o} & 38.5 & 40.2 & 43.2 & 49.5 \\
\small{o1-mini} & 25.0 & 35.0 & 65.0 & 60.1 \\
\small{o1-preview*} & 45.0 & 26.8 & 70.1 & 50.2 \\
\small{claude-3.5-sonnet} & 39.1 & 42.0 & 50.4 & 46.0 \\
\small{claude-3-opus} & 32.2 & 41.7 & 49.4 & 47.0 \\
\small{gemini-1.5-pro} & 30.9 & 41.5 & 38.1 & 40.5 \\
\small{gemini-1.5-flash} & 35.5 & 35.7 & 38.3 & 45.5 \\
\small{Ours} & 40.0 & 56.1 & 57.2 & 54.8 \\
\bottomrule
\end{tabularx}
\caption{Experimental results on 3 candidate moves.}
\label{table2}
\end{table}

Since our data collection pipeline is automatic, we are able to add more reasonable candidate moves for a chess board position to our dataset conveniently. We conduct additional experiments given chess positions with 3 candidate moves. We sample 1000 positions from the test set of MATE for our new test sets; for each position, we sample 3 candidate moves and then annotate them. We evaluate models on 1000 samples in the new test sets. As we point out, prompting strategies do not significantly impact performance in our chess task(in Section 4.2), we use the zero-shot setting. We combine the evaluation results of our four finetuned models as 'Ours' in the Table \ref{table2}.

With increasing numbers of candidate moves, we observe a decline in model performance. Notably, models finetuned with strategy and tactical explanations demonstrate greater robustness when adapting to novel and more challenging tasks, compared to models finetuned without such explanations.

\subsection{Experiments on Generating Explanations}\label{appendix:generate}

\begin{table}[htbp]
\centering
\begin{tabular}{l@{\hspace{8pt}}c@{\hspace{8pt}}c@{\hspace{8pt}}c}
\toprule
& MATE-gpt & MATE-claude & MATE-ours \\
\midrule
gpt & -- & 48.6 & \textbf{51.0} \\
claude & 52.7 & -- & \textbf{56.7} \\
ours & \textbf{74.7} & \textbf{75.6} & -- \\
\bottomrule
\end{tabular}
\caption{Evaluating models' capability to generate strategic explanations.}
\label{table3}
\end{table}

We conduct experiments to evaluate models' capability of generating strategy explanations. We fintune our models using the pretrained llama-3-8B model as the foundation model. The training set and the test set are modified from MATE: for each sample, the input takes the chess board position and move, the output is the strategy explanation or tactic explanation. During training, we employ a cosine learning rate scheduler with 3\% warm-up steps. The maximum learning rate is $5\times 10^{-6}$. We train the model over 8$\times$H100 GPU for 10 epochs.

We modify the test set for measuring models' strategy generation. To measure our model's generated explanations, we sample 1000 positions with candidate moves, instead of following our data anotaion process, we use our model to generate strategy explanations for the test set MATE-ours. Similarly, for the same 1000 positions and candidate moves, we use gpt-4o to generate strategy explanation for the test set MATE-gpt. We craft test set MATE-claude using claude-3.5-sonnet. We test gpt-4o, claude-3.5-sonnet, and our model's chess playing by choosing the right move given a position and two candidate moves in the test set MATE-ours, MATE-gpt, MATE-claude respecitively. The experiments results are shown in Table \ref{table3}.

Based on the performance across these test sets, we find that our model's strategy generation are better compared with gpt-4o claude-3.5-sonnet. The experiments demonstrate the our model's intrisic reasoning capability outperform those commercial models in chess.

\subsection{Difficulty Levels of Sub-Datasets}\label{appendix:difficulty}
Our MATE consists of 4 sub-datasets: MATE-N, MATE-S, MATE-T, and MATE-ST. We conduct two experiments to study the difficulty levels of chess board positions across all these sub-datasets through both human and automatic assessment.

\begin{table}[htbp]
\centering
\begin{tabularx}{\linewidth}{lXXXX}
\toprule
\textbf{Model} & \textbf{N} & \textbf{S} & \textbf{T} & \textbf{ST} \\
\midrule
gpt-4o & 46.4 & 47.4 & 46.0 & 46.5 \\
claude-3.5-sonnet & 49.6 & 51.2 & 50.2 & 48.6 \\
\bottomrule
\end{tabularx}
\caption{Experimental results in terms of accuracy(\%) on 1000 board positions selected from MATE-N, MATE-S, MATE-T, MATE-ST.}
\label{table4}
\end{table} 

We first conduct an experiment with chess players. From each sub-dataset, we randomly select 50 samples, retaining only the board position and candidate moves while omitting any strategy or tactical information. Players are then asked to rate the difficulty of these samples. The results indicate that human players perceive the positions and candidate moves in all four sub-datasets to be of similar difficulty levels.

For our second experiment, we employ state-of-the-art commercial large language models to assess the difficulty levels of the sub-datasets. We randomly selected 1000 samples from each sub-dataset, preserving only the board position and candidate moves while excluding any strategic or tactical information. The language models were then prompted to determine the optimal move for each position.
We utilized gpt-4o and claude-3.5-sonnet for this experiment. The results, presented in Table \ref{table2}, indicate that these language models performed similarly across the samples selected from all sub-datasets, suggesting the same difficulty levels of these sub-datasets.
\subsection{Case Study}
\begin{figure*}[t!]
    \centering
    \includegraphics[width=\linewidth,trim=0cm 0cm 0cm 0cm,clip]{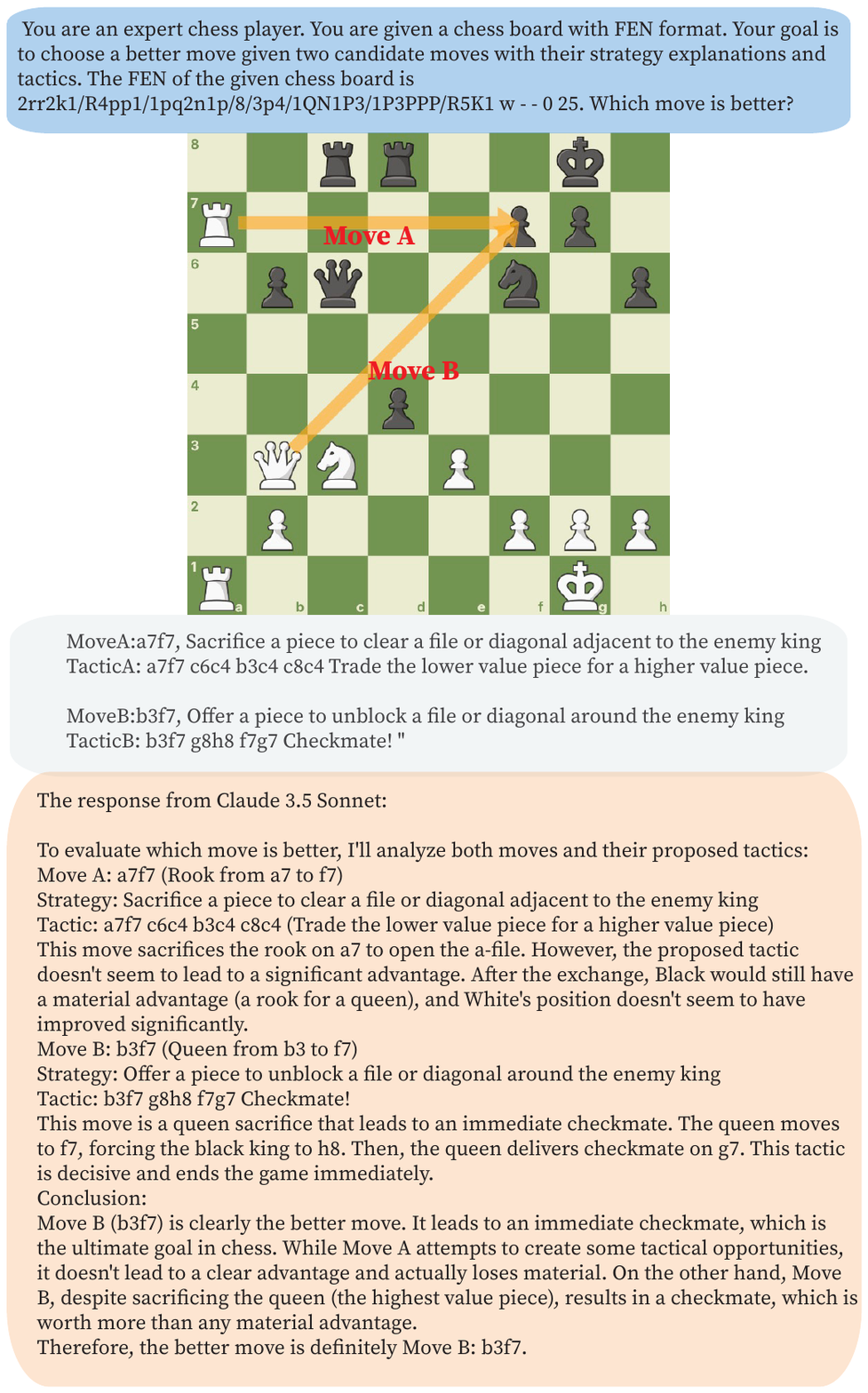}
    \caption{\textbf{Case Study:Claude 3.5 Sonnet.}}   
    \label{fig:casestudy_claude}
\end{figure*}

\begin{figure*}[t!]
    \centering
    \includegraphics[width=\linewidth,trim=0cm 0cm 0cm 0cm,clip]{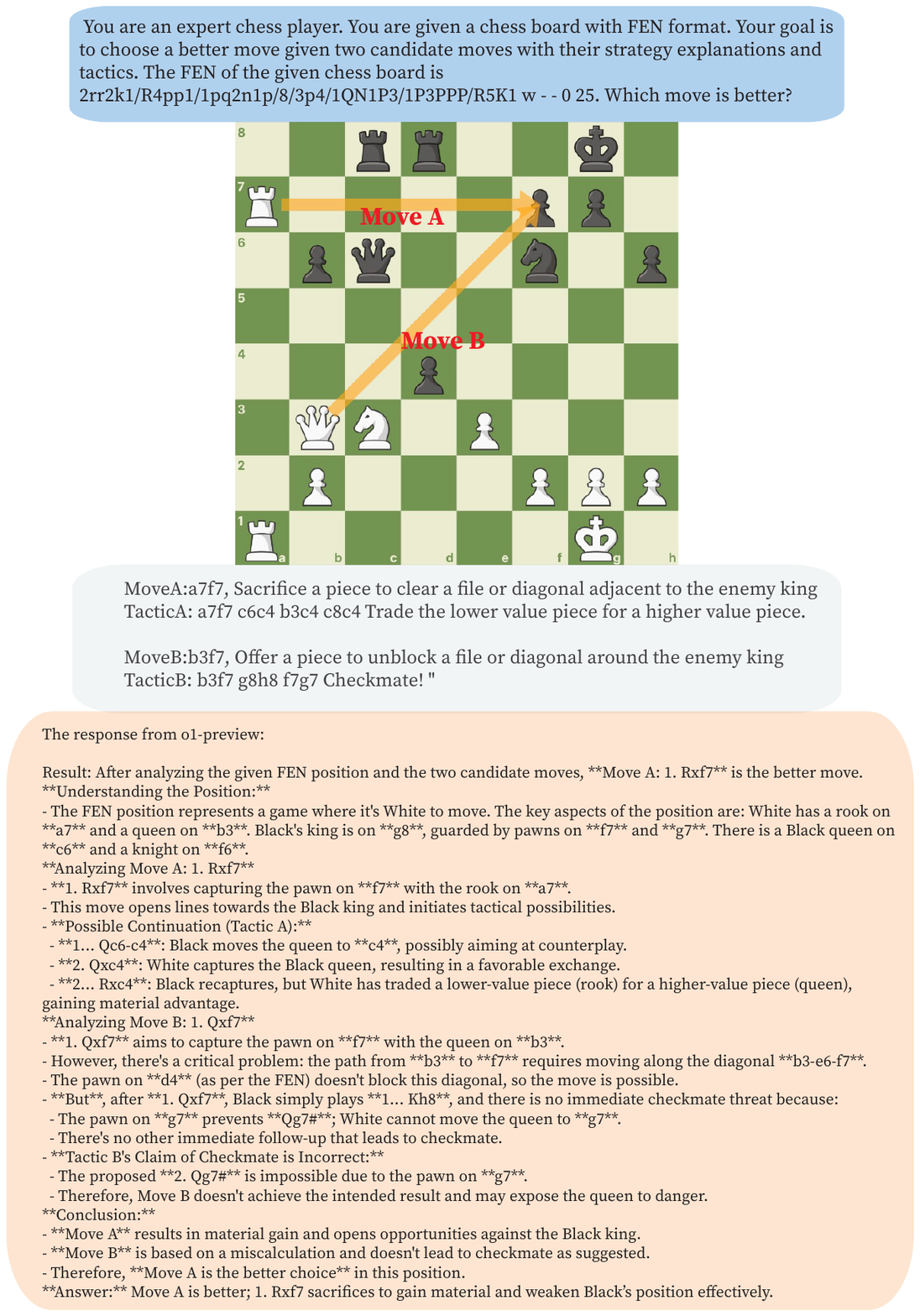}
    \caption{\textbf{Case Study:o1-preview.}}   
    \label{fig:casestudy_o1}
\end{figure*}

\begin{figure*}[t!]
    \centering
    \includegraphics[width=\linewidth,trim=0cm 0cm 0cm 0cm,clip]{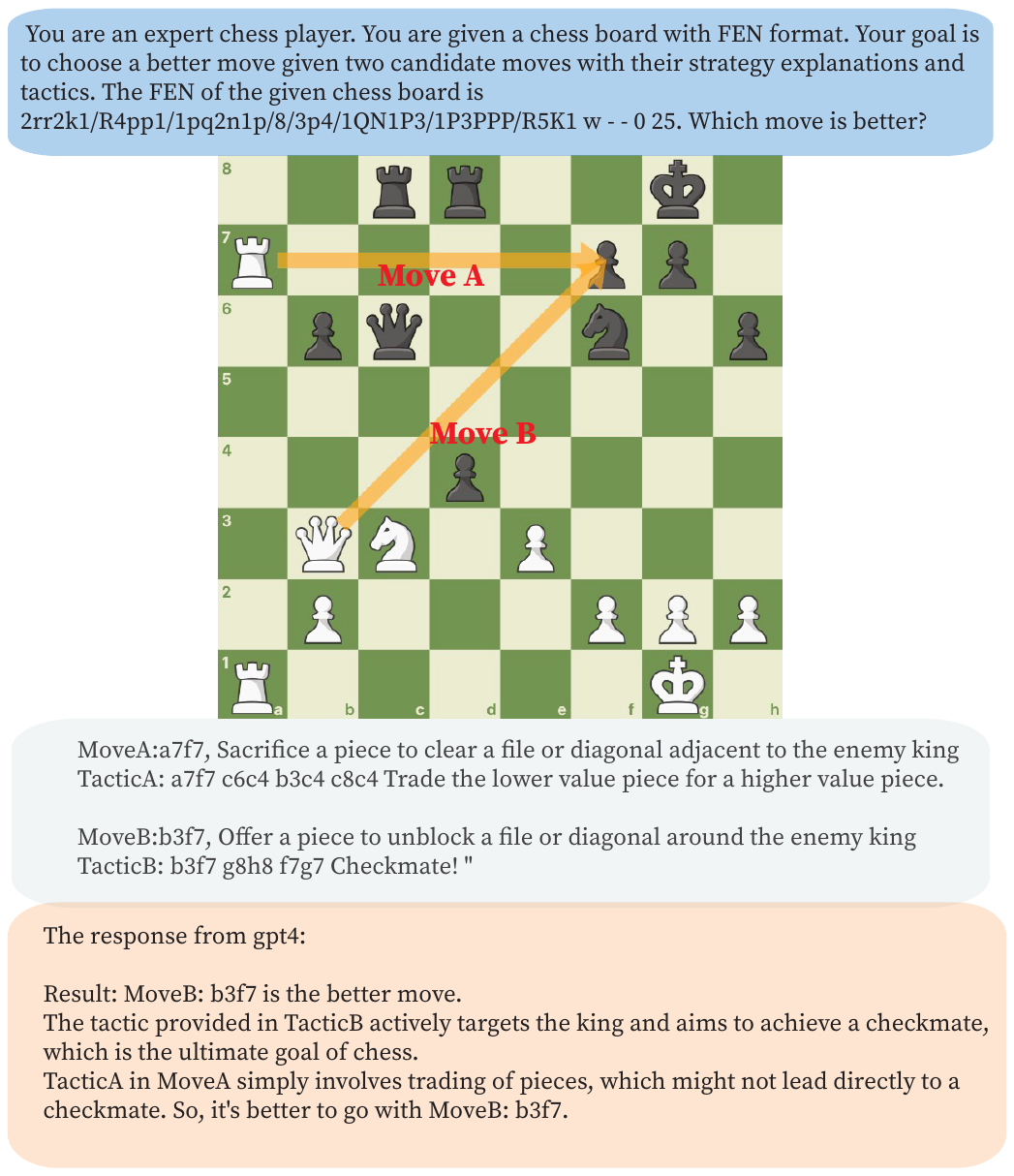}
    \caption{\textbf{Case Study:gpt-4.}}   
    \label{fig:casestudy_gpt4}
\end{figure*}

We pick a sample case with both strategy and tactic annotated, and show the responses from three language models. See Figure\ref{fig:casestudy_claude}, Figure\ref{fig:casestudy_o1}, and Figure\ref{fig:casestudy_gpt4}.

\end{document}